\documentclass[10pt,twocolumn,letterpaper]{article}

\usepackage{wacv}
\usepackage{times}
\usepackage{epsfig}
\usepackage{graphicx}
\usepackage{amsmath}
\usepackage{amssymb}

\usepackage[ruled,linesnumbered,vlined]{algorithm2e}

\wacvfinalcopy 

\def\wacvPaperID{54} 

\ifwacvfinal\fi
\begin{document}

\title{Instance-based Deep Transfer Learning}

\author{Tianyang Wang \\
Austin Peay State University\\
{\tt\small toseattle@@siu.edu}
\and
Jun Huan \\
Baidu Research\\
{\tt\small huanjun@baidu.com}
\and
Michelle Zhu \\
Montclair State University\\
{\tt\small zhumi@montclair.edu}
}

\maketitle
\ifwacvfinal\thispagestyle{empty}\fi

\begin{abstract}
   Deep transfer learning recently has acquired significant research interest. It makes use of pre-trained models that are learned from a source domain, and utilizes these models for the tasks in a target domain. Model-based deep transfer learning is probably the most frequently used method. However, very little research work has been devoted to enhancing deep transfer learning by focusing on the influence of data. In this paper, we propose an instance-based approach to improve deep transfer learning in a target domain. Specifically, we choose a pre-trained model from a source domain and apply this model to estimate the influence of training samples in a target domain. Then we optimize the training data of the target domain by removing the training samples that will lower the performance of the pre-trained model. We later either fine-tune the pre-trained model with the optimized training data in the target domain, or build a new model which is initialized partially based on the pre-trained model, and fine-tune it with the optimized training data in the target domain. Using this approach, transfer learning can help deep learning models to capture more useful features. Extensive experiments demonstrate the effectiveness of our approach on boosting the quality of deep learning models for some common computer vision tasks, such as image classification.
\end{abstract}

\section{Introduction}

Transfer learning has been a classical problem in machine learning. There are four main strategies of implementations \cite{pan2010survey}: instance-based, feature-based, model-based, and relation-based. Recently, deep transfer learning has attracted a lot of research attentions, and many efforts have been made to bridge the gap between the source and target domains. Typical methods include \cite{tzeng2017adversarial, ge2017borrowing, pmlr-v70-long17a}, which focus on domain adaptation by feature alignment. However, these methods are more suitable for unsupervised or semi-supervised deep transfer learning. As indicated in \cite{zagoruyko2016paying}, under fully supervised context, finding a way to boost up the quality for an existing model is highly desired than training from scratch in a target domain in practice.  

Moreover, in deep learning, a popular approach is to
fine-tune pre-trained models, that are learned from large benchmark datasets in source domains, to achieve ideal results in other similar target domains. As a typical model-based implementation, people can choose to tune the entire models or only fine-tune several deep layers since training a model from scratch is time-consuming. 
Although the model-based approach is effective, very little progress has been made on exploring how individual data in a target domain will influence the learning performance in this domain. 
In the traditional transfer learning, instance-based approaches need to measure the similarity between a source and a target domain, and adjust the weight values for the source domain samples which have similar samples in the target domain \cite{pan2010survey}. However, this strategy is not suitable for deep learning models due to two main reasons. Firstly, the estimation of similarity highly relies on method and experience, thus it is challenging to be integrated into an end-to-end learning fashion. Secondly, determining appropriate weight values is still difficult and subjective. Furthermore, such a strategy lacks flexibility and highly relies on fine-tuning skills.

To enhance the performance of pre-trained models in a target domain and explore a
feasible combination of transfer learning and deep supervised learning, we propose an instance-based deep transfer learning approach. Specifically, given a target domain, we first select a similar source domain which has much more training data than the target domain. We choose a pre-trained model that was learned from the source domain, and use this model to estimate the influence of each training sample in the target domain. The influence value measures how an individual training sample will impact the performance of the pre-trained model in the target domain. According to the influence, we can determine to preserve or remove a training sample in the target domain. After removing several training samples, we obtain an optimized training set which consists of fewer samples than the original training set. 
This optimization can implicitly filter out disturbing features which will jeopardize the model performance. Besides, it can also alleviate the imbalanced classification problem, in which most samples are from only one class. After the training data optimization in a target domain, we can fine-tune the pre-trained model with the optimized training data or create a new model and initialize it based on the pre-trained model. According to the target domain tasks, we perform either global or local fine-tuning to obtain a final predictive model. We validate our approach on widely used datasets, and the experimental results demonstrate its effectiveness on boosting deep learning models. In addition, we describe an uncommon but possible data distribution in a target domain, and show that our approach is also capable of handling such a task. 

Even though our approach does not explicitly analyze the relationship between a source and a target domain, it can still effectively help deep learning models to learn more useful features. This can be attributed to two main reasons. Firstly, choosing a pre-trained model from a similar source domain implicitly incorporates data relevance into the target domain learning. Secondly, training data optimization implicitly removes the disturbing features. When a new model is created for the target domain, we at least need to initialize several shallow layers by using the corresponding layers of the pre-trained model, because shallow layers of the pre-trained model learned significant global features from the source domain \cite{yosinski2014transferable}, and these features are also critical for the target domain learning. It is worth noting that, while our approach also exploits pre-trained models, it is essentially different from the model-based deep transfer learning since we utilize pre-trained models to estimate data influence in a target domain in addition to loading parameters, whereas typical model-based approaches only use pre-trained models for fine-tuning. 

The main contributions of this paper are two folds. First of all, we propose an instance-based deep transfer learning approach to boost model performance in a target domain. Secondly, we provide a different perspective for analyzing how transfer learning can be integrated with deep learning. 

\section{Related Work}
\label{related}
In this section, we review literature from two aspects, namely training data optimization, and image classification models that will be used in our experiments.


\subsection{Training Data Optimization}
\label{opt}
Although a lot of research works have been focusing on designing powerful network models to boost learning performance, very little progress has been made on investigating the influence of training data on learning effects. Koh and Liang in \cite{pmlr-v70-koh17a} pioneer to study how a single training sample will influence the testing loss of a single testing sample. Inspired by their work, Wang et al. in \cite{wang2018data} propose a scheme to optimize training data, namely \emph{data dropout}. In our work, we modify their scheme to optimize the training data in a target domain, aiming to implement our instance-based deep transfer learning approach. To ease the discussion, we use $\pmb{\mathcal{A}}$ to denote this modified version, which is detailed in the \textbf{Algorithm 1}. 
$x_{i}$ ($i=1,...,n$) denotes a single training sample, and $f_{\theta}(x)$ a model such as CNN with input $x$. Let $L(f_{\theta}(x))$ be the loss, and $I_{loss}(x, x_{j})$ the influence of removing a training sample $x$ on the loss at a validation sample $x_{j}$, where $j=1,...,k$ and $k$ equals the number of validation samples. If validation data is not given in advance, one can build a validation set by randomly separating 10\% of the total samples from the given training set. The general goal of training a model aims to learn a set of parameters $\theta = \arg\min_{\theta} \frac{1}{n} \sum_{i=1}^n L(f_{\theta}(x_{i}))$, where $L$ can be typical loss functions used in deep learning, such as softmax loss, mean squared error (MSE), and $L_2$ loss, etc.

\begin{algorithm}[]
\SetKwData{P_u}{P_u}\SetKwData{P_v}{P_v}\SetKwData{P_i}{P_i}\SetKwData{P_j}{P_j}
\SetKwData{T1}{t1}\SetKwData{T2}{t2} \SetKwData{E}{E}
\SetKwData{Left}{left}\SetKwData{This}{this}\SetKwData{Up}{up}
\SetKwInOut{Input}{Input}\SetKwInOut{Output}{Output}
\Input{\\
$f_{\theta}$: pre-trained network from source domain \; 
$\mathcal{X}$: training set in target domain \;
$\mathcal{V}$: validation set in target domain\; 
$i$/$j$: training/validation sample index\;
$len$($\cdot$): number of samples in a dataset ``$\cdot$"\;}
\Output{\\
$\mathcal{X'}$: optimized training set in target domain\;
}
\Begin{
\For{$i\leftarrow 1$ \KwTo $len$($\mathcal{X}$)}{
\For{$j\leftarrow 1$ \KwTo $len$($\mathcal{V}$)}{ use $f_{\theta}$ to
\textbf{compute} $I_{loss}(\mathcal{X}(i)$, $\mathcal{V}(j))$;
}
\If{$\sum_{j}I_{loss}(\mathcal{X}(i)$, $\mathcal{V}(j))>0$}
   {\textbf{remove} $\mathcal{X}(i)$ from $\mathcal{X}$;}
}   
$\mathcal{X'}$ is obtained\;
}
\caption{Algorithm $\pmb{\mathcal{A}}$}
\label{algo_iterative}
\end{algorithm} 

According to \cite{pmlr-v70-koh17a}, 
$I_{loss}(x, x_{j})$ can be approximated by 
\begin{equation}
\label{eq3}
  I_{loss}(x, x_{j})=-\nabla_\theta L(f_{\theta}(x_{j}))^\top H_{\theta}^{-1} \nabla_\theta L(f_{\theta}(x)),
\end{equation}
where $H_{\theta}^{-1}=\frac{1}{n} \sum_{i=1}^n \nabla^2_\theta L(f_{\theta}(x_{i}))$ is the Hessian and positive definite. For each training sample $x$, its influence on the loss across all validation samples can be computed by $\sum_{j}I_{loss}(x, x_{j})$. This is the value that we want to measure to reflect how the validation loss will vary if a single training sample $x$ is removed from the given training set. In fact, $\sum_{j}I_{loss}(x, x_{j})$ is equivalent to 
\begin{equation}
\label{eq4}
  \sum_{j} L(f_{\theta}(x_{j}))-L(f_{\theta'}(x_{j})), 
\end{equation}
where $\theta' = \arg\min_{\theta} \frac{1}{n} \sum_{x_{i}\neq{x}} L(f_{\theta}(x_{i}))$. One will expect $\sum_{j} L(f_{\theta}(x_{j}))-L(f_{\theta'}(x_{j}))>0$, which implies that removing a training sample $x$ can decrease the total validation error. Since its equivalent item is $\sum_{j}I_{loss}(x, x_{j})$, the criteria of removing a training sample $x$ from the given training set can be set to: $\forall{x}$, if $\sum_{j}I_{loss}(x, x_{j})>0$, $x$ will be dropped from the given training set, otherwise, $x$ will be preserved \cite{wang2018data}. 
The algorithm in \textbf{$\pmb{\mathcal{A}}$} describes how to optimize the training set in a target domain. $\mathcal{X'}$ is a reconstructed training set which consists of fewer samples than the original training set. In step 4 based on eq.\ref{eq3}, we need a pre-trained network to compute the influence value, and pre-trained models from a relevant source domain are used for this purpose.

\subsection{Models for Image Classification}
\label{models}
Designing powerful neural networks for effective image classification is a major interest in deep learning research. Typical models include VGG network \cite{simonyan2014very}, ResNet \cite{he2016deep}, and DenseNet \cite{huang2017densely}, which presents the state-of-the-art results. Compared to the VGG network which is actually a sequential model, ResNet and DenseNet creatively adopt skip connections to learn more distinguished features by fusing the outputs of different layers. The pre-trained models can be found easily in research community and we in this work will use them directly to perform training data optimization in a target domain, network initialization for a new created model, as well as direct fine-tuning if unnecessary to build a new model.

\section{Approach}
\label{Method}

Given a pre-trained model $f_{\theta}$ that was learned from a source domain, the goal of the model-based deep transfer learning can be expressed as follows, 
\begin{equation}
  f_{\theta} \xrightarrow{X} f_{\theta'},
\end{equation}
where $\theta$ denotes the pre-trained parameters and $\theta'$ the parameters learned from the dataset $X$ in a target domain. Instead of learning from $X$, our instance-based approach will reconstruct an optimized training set, and the learning process can be described as 
\begin{equation}
  opt(X) \xrightarrow{} X', \quad f_{\theta} \xrightarrow{X'} f_{\theta'},
\end{equation}
where $opt$ denotes the training data optimization that is performed based on the algorithm $\pmb{\mathcal{A}}$, and $X'$ is the optimized training set in a target domain. 

Next, we will briefly clarify the term \emph{instance} and introduce the data dropout scheme \cite{wang2018data}. Then we propose our instance-based deep transfer learning approach, and give some useful discussions relevant to its usage. Moreover, we also describe an uncommon but possible data distribution in a target domain that we will utilize to validate our approach in the experiments section.

\subsection{Instance and Data Dropout}
Since our work focuses on transfer learning, it naturally involves two groups of data from a source and a target domain, respectively. The term \emph{instance} that we use in the context only refers to the training samples in the target domain. This is because our approach will not operate the source domain, and the only connection between the source and target domain is a pre-trained model that is learned from the source domain. We employ this model to evaluate each \emph{instance} in the target domain.

Data dropout aims to identify bad training samples which will jeopardize the model performance. Thus these \emph{negative} samples should be removed. 
Wang et al. in \cite{wang2018data} provide a practical scheme to remove negative training samples, and it has been proven effective for unstructured data such as images. In our work, we use this scheme to optimize the training data in a target domain. For a specific training sample in a target domain, if the computed influence value is positive, it will be removed. Such an influence value is evaluated over the validation data rather than testing data, since testing data should be invisible during training phase and usually not available until testing phase. Detailed discussion of the data dropout scheme is beyond the scope of this work, and readers can refer to our modified version in section \ref{opt}.  

\subsection{Instance-based Deep Transfer Learning}
In typical model-based deep transfer learning, pre-trained models that were learned from a source domain are adopted and fine-tuned to fit the data in a target domain. Unlike this conventional manner, our approach adds a simple yet effective step to measure the influence of each individual training sample in a target domain. The main purpose is to improve model performance in the target domain by optimizing the training data in the target domain. 

Specifically, given a target domain, we choose a pre-trained model from a similar source domain. The selected source domain should have much more data, and have similar feature space as the target domain. For instance, if the target domain contains the two CIFAR datasets \cite{krizhevsky2009learning}, we can choose a pre-trained model from the ImageNet dataset \cite{deng2009imagenet}. Once a pre-trained model is selected, we use it to estimate the influence of each training sample in the target domain. According to the influence value, we will remove several negative training samples by following the algorithm $\pmb{\mathcal{A}}$, hence the training set in the target domain is actually optimized. Afterwards, we continue to fine-tune the pre-trained model or build a new model and train it with the optimized training set. For a pre-trained model, initialization is not a concern since its parameters are learned from a similar source domain. For a new created model, it can be initialized in a hybrid way: shallow layers inherit parameters from the pre-trained model and deep layers are initialized by random parameters. We generalize the steps in the \textbf{Approach}. When fine-tuning $f_{\theta_{pre}}$ in step 3, we may need to adjust the pooling layers. This is due to the fact that the ImageNet pre-trained models were learned from images of large dimension, and several pooling layers need to be removed when fine-tuning on smaller size images. $f^{+}_{\theta_{pre} \& \theta_{0}}$ means the model $f^{+}$ is initialized in a hybrid way of using both pre-trained and randomized parameters. The experiment in section \ref{local} is a typical example of using hybrid initialization.

Note that in \cite{wang2018data}, the authors also utilize a model to compute data influence, however, our approach distinguishes from their work in the following aspects. Firstly, it does not involve two domains since transfer learning is not considered in their work. While in our paper, we mainly consider the scenario of deep transfer learning. Secondly, after the training data optimization, they re-train the model from scratch by re-initializing the network with randomized parameters. While in our approach, we continue to fine-tune a pre-trained model with an optimized training set. Although we may also build a new model according to the task in a target domain, the initialization for the created model cannot be independent of the pre-trained model, which we use to estimate the influence of each training sample in the target domain. Therefore, data relevance between the source and target domain is always implicitly considered in our work.


\begin{table}
\normalsize
\centering
\vspace{-1mm}
\begin{tabular}{l}
\hline
\textbf{Approach}: Instance-based Deep Transfer Learning
\\
\hline
Pre-Trained Model: $f_{\theta_{pre}}$,  \\
Training Set in Target Domain: $\mathcal{X}$; \\
\textbf{Optimized} Training Set in Target Domain: $\mathcal{X'}$; \\ 
Created Network: $f^+$;  \\
Randomized Parameters: $\theta_{0}$;

\\
\\
1. Optimize $\mathcal{X}$ to obtain $\mathcal{X'}$ based on algorithm $\pmb{\mathcal{A}}$; 
\\
2. Initialize $f^+$ with $\theta_{pre} \& \theta_{0}$ if $f^+$ is created;
\\
3. Fine-tune $f_{\theta_{pre}}$ or $f^{+}_{\theta_{pre} \& \theta_{0}}$ with $\mathcal{X'}$;
\\
\hline
\end{tabular}
\vspace{-3mm}
\label{a1}
\end{table}

\subsection{Discussions}
After training data is optimized in a target domain, there are two options: continue to fine-tune a pre-trained model, or create a new model and initialize it partially based on the pre-trained model. For instance, if we only want to load pre-trained parameters for shallow layers, we can create a new model and initialize other layers with randomized parameters. In general, the decision can be made according to the data in a target domain, and the experiment in section \ref{local} gives such an example. In fact, if a new model is created, shallow layers should be at least initialized by pre-trained parameters. Otherwise, the estimation of data influence in prior steps will not be reliable for fine-tuning.

We utilize global fine-tuning strategy in our work, thus all parameterized layers in a pre-trained model will be fine-tuned. Unlike freezing several layers, global fine-tuning can further boost network performance as validated in \cite{yosinski2014transferable}. Moreover, it has been proven in \cite{mcnamara2017risk} that global fine-tuning has a lower risk bound in transfer learning. 

Throughout the analysis, we assume that a target domain and a selected source domain have similar feature space. This is because we use a pre-trained model from the source domain to compute the influence of each training sample in the target domain. The results will not be reliable if the source and target domain are quite different. Therefore, our approach belongs to the category of homogeneous transfer learning \cite{pan2010survey}. On the other hand, as our approach is independent of models and domains, it can be easily extended for different applications.

\section{Experiments}
\label{experiment}

To validate the effectiveness of our approach, we conduct extensive experiments on image classification problems. Besides the widely used data sets which follow uniform or near uniform distribution, we also investigate an unusual but possible data distribution, and apply our approach to classify such data. It is worth noting that we need a validation set to estimate the influence of each training sample in a target domain, and we will randomly separate 10\% of the training samples in the target domain to build a validation set if it is not given. Unless specified, we will use this way to build a validation set when validation samples are needed.   

\subsection{A Warm-up Example}
With this experiment, we aim to validate the feasibility of combining transfer learning and the algorithm $\pmb{\mathcal{A}}$. Such a combination is the foundation of our instance-based deep transfer learning. 
We conduct the experiment using two datasets, namely EMNIST-Letters \cite{cohen2017emnist}, and MNIST \cite{lecun1998mnist}. The latter one is widely used as an entry-level dataset for evaluating classification algorithms. It contains 10 classes of handwritten digits (eg, 0, 1, ..., 9), and each digit is wrapped in an image of dimension $28 \times 28$. There are 60000 training images and 10000 testing images.  EMNIST-Letters dataset is recently proposed, and it contains 37 classes of handwritten letters (eg, a, b, ...z, and their variations). Each letter is also wrapped in an image of dimension $28 \times 28$ which has the same structure as the original MNIST dataset. There are 88800 training images and 14800 testing images. 

The classification task on MNIST is simple and training a convolutional neural network (CNN) from scratch will not be computing-intensive, therefore, it can hardly find a pre-trained model in research community. Although the handwritten letters look different from the digits, the strokes that compose the letters and the digits are similar. Thus the two datasets can be regarded as sharing similar feature space. As a result, we train a CNN model on the EMNIST-Letters dataset first and use it as a pre-trained model for the MNIST task. 
Specifically, we create a model with 4 convolutional layers with the ReLU \cite{nair2010rectified} activation and two max-pooling layers. Two fully connected layers are linked to a softmax layer which is used to compute the classification probability. We initialize the model using the \emph{Xavier} method \cite{glorot2010understanding}, and train it using the Adam optimizer \cite{kinga2015method}. Default parameters of the Adam optimizer are not changed, such as the learning rate of 0.001. We set a mini-batch size of 128 and train this model on the EMNIST-Letters dataset for two epochs. Then we stop the training and use the resulting model as a pre-trained model to estimate the influence of each training sample in the MNIST dataset, and optimize the training set by following the algorithm $\pmb{\mathcal{A}}$. Then we fine-tune the pre-trained model with the optimized training set for only one epoch. This is because the MNIST dataset is easy to be fitted, and an arbitrary deep CNN model can achieve a testing accuracy of above 99\% if trained for a few epochs. In that case, it cannot differentiate whether it is the contribution of our approach. To have reasonable comparison, we also train the same model from scratch on the MNIST dataset for one epoch. In addition, we directly fine-tune the pre-trained model without using the optimized training set. Such an implementation is a typical example of the model-based deep transfer learning.

\renewcommand{\arraystretch}{1.06}
\setlength{\tabcolsep}{1.0em}
\begin{table}[]
\tiny 
\resizebox{0.45\textwidth}{!}{%
\begin{tabular}{c|l|c}
\hline
\multicolumn{1}{c|}{Strategy} & Model & MNIST  \\ \hline

from Scratch  & & 1.37  \\ 

Model-based TL & sample-CNN  & 1.09  \\ 

Instance-based TL  & & 0.47  \\ \hline
\end{tabular}%
}
\vspace{0.5ex}
\caption{Testing error rates (\%) of the sample model on the MNIST classification.}
\label{4-1table}
\vspace{-3mm}
\end{table}

We compare the results of the three training schemes in Table \ref{4-1table}. Model performance is measured in testing error rate. As can be seen, the model trained with our instance-based approach achieves a lower error rate than the model trained with the other two methods. Therefore, the optimized training set can boost the performance of the pre-trained model in the target domain. This indicates that the transfer learning is compatible with the algorithm $\pmb{\mathcal{A}}$. In the following experiments, we will utilize this algorithm to optimize the training data in a target domain.

\subsection{Classification of Real Color Images}
\label{cifarclassify}
To validate the effectiveness of our approach, we conduct experiments on several widely used datasets for image classification, including the two CIFAR datasets \cite{krizhevsky2009learning}, and the SVHN (Street View House Number) dataset \cite{netzer2011reading}. The CIFAR-10 dataset contains 10 classes while the CIFAR-100 has 100 classes of real color images. Each image has a dimension of $32 \times 32$. Each dataset contains 50000 training images and 10000 testing images, which are uniformly distributed in 10 and 100 classes. The SVHN dataset contains 10 classes of 73,257 training images and 26,032 testing images. It also has an additional training set having 531,131 images for further training. 

We implement our approach based on three well-known deep CNN models, namely VGG-16 \cite{simonyan2014very}, ResNet-50 \cite{he2016deep} and DenseNet-121 \cite{huang2017densely}. As mentioned in section \ref{models}, these models can learn distinguished features from real color images, hence are widely used in image classification tasks. Detailed discussion of the architectures is beyond the scope of this work, interested readers can refer to the original papers for details.
To perform instance-based deep transfer learning, we choose the pre-trained versions, that were learned from the ISLVRC dataset \cite{deng2009imagenet}. Such pre-trained models are usually named as \emph{ImageNet pre-trained} models. The ISLVRC dataset is a subset of the ImageNet dataset \cite{deng2009imagenet}, and it has around 1.28 million images for training, 50000 images for validation. There are 100000 testing images uniformly distributed in 1000 classes. In practice, all these color images can be cropped to a fixed size, such as $224 \times 224$. In this work, we choose the ImageNet pre-trained models mainly due to two reasons. Firstly, the ImageNet images are similar to the images in the two CIFAR datasets, hence they have similar feature space. Secondly, ImageNet pre-trained models are supported by well-known deep learning platforms, such as Tensorflow \cite{tensorflow2015-whitepaper} and Keras \cite{chollet2015keras}. 
We use the pre-trained models to compute the influence of each training sample in a target domain by following the algorithm $\pmb{\mathcal{A}}$. For the two CIFAR datasets, we know the testing data in advance. However, in real scenarios, testing data may be unavailable until testing phase. Therefore, we use the validation data in a target domain as a reference to estimate the influence of the training data in the target domain. 

It is important to note that the pre-trained models were trained with large dimensional images, such as $224\times224$ for the ResNet-50. However, when  fine-tuning the models on smaller dimensional images, such as $32\times32$ of the CIFAR data, there will be an issue of incompatible dimension, we thus adjust pooling layers accordingly. Take the ResNet-50 as an example, we remove the two pooling layers which are not in the convolutional blocks. In addition, the pre-trained models provided by the well-known platforms usually have a limit of input size. To avoid issues, we create a duplicate model and load pre-trained parameters manually. 
We use the Adam optimizer to perform fine-tuning because its default parameters can be used for most models and applications. Moreover, since we need to fine-tune multiple models, and also need to train multiple models from scratch for the purpose of comparison, changing hyper-parameters frequently may result in unpredictable issues. Typical data augmentation methods, such as the horizontal flipping and translation are utilized during training. For the pre-trained VGG-16 and ResNet-50 models, we fine-tune them for 100 epochs, while the DenseNet-121 is fine-tuned for 150 epochs. We multiply the default learning rate of 0.001 with a fixed factor of 0.1 at the 50\% of the total epochs in each fine-tuning process. A mini-batch size of 128 is used for the VGG-16 and ResNet-50, while 64 for the DenseNet-121. In addition, to compare our approach with the traditional training method which is not based on transfer learning, we also train the selected models from scratch without using our approach. In this case, we randomly initialize the networks instead of inheriting weights from the pre-trained models. Both VGG-16 and ResNet-50 are trained for 200 epochs, whereas the DenseNet-121 is trained for 300 epochs.

\renewcommand{\arraystretch}{1.06}
\setlength{\tabcolsep}{1.0em}
\begin{table}[]
\resizebox{0.45\textwidth}{!}{%
\begin{tabular}{c|l|c|c}
\hline
\multicolumn{1}{c|}{Strategy} & Model & CIFAR-10 & CIFAR-100 \\ \hline
  & VGG-16  & 7.81 & 28.82 \\
from Scratch & ResNet-50  & 6.93 & 27.59 \\ 
  & DenseNet-121 & 4.98 & 21.19  \\\hline

  & VGG-16  & 7.14 & 28.01 \\
Model-based TL & ResNet-50  & 6.20 & 26.90 \\ 
  & DenseNet-121 & 4.23 & 20.72  \\\hline 

  & VGG-16  & 5.01 & 25.29 \\
Instance-based TL & ResNet-50  & 4.11 & 24.17 \\ 
  & DenseNet-121 & 2.45 & 18.67  \\\hline
\end{tabular}%
}
\vspace{0.5ex}
\caption{Testing error rates (\%) of the selected models, which are trained by different approaches, on the CIFAR classification.}
\label{4-2-1table}
\end{table}

\begin{figure}
    \centering
    \includegraphics[width=0.2\textwidth]{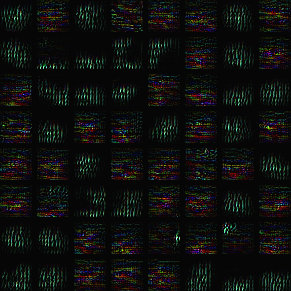}
    \includegraphics[width=0.2\textwidth]{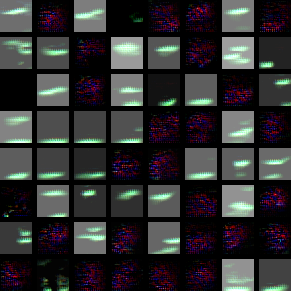}
    \caption{For a test image which is misclassified by model-based approach while correctly classified by instance-based approach, the learned feature maps in the 23rd convolutional layer of the ResNet-50. \emph{Left:} trained by model-based approach; \emph{Right:} trained by instance-based approach.}
    \vspace{-3mm}
    \label{filters}
\end{figure}

We compare three groups of results in Table \ref{4-2-1table}. In the first group, the selected models are trained from scratch and their parameters are randomly initialized by the \emph{Xavier} method \cite{glorot2010understanding}. In the second group, the selected models were pre-trained on the ImageNet data, and we fine-tune them without optimizing the training data in the target domain. In the third group, we still use the same pre-trained models, but optimize the training data in the target domain by following the algorithm $\pmb{\mathcal{A}}$. The third group is the result of our instance-based deep transfer learning approach, while the second group corresponds to the model-based deep transfer learning. 
The results are measured in testing error rates of classification. As can be seen, the second group of results are better than the first group, and our approach further improves the results of the second group. This indicates that our instance-based learning is more advantageous than the model-based method. By optimizing the training data in a target domain, more significant features can be learned while disturbing features are removed. Here, we perform global fine-tuning for all layers to achieve better results. For comparison, we show the effects of local fine-tuning for the ResNet-50 in Table \ref{4-2-2table}, where several pre-trained layers are frozen and the model is fine-tuned for 90 epochs. It can be seen that the results are inferior to that in Table \ref{4-2-1table}. Therefore, if a target domain has sufficient training data, fine-tuning an entire model is preferred. It is interesting to note that the selected pre-trained models may have inferior performance than their shallower counterparts. For example, training the ResNet-20 from scratch on the CIFAR data can give even better results than ours. This is because the selected pre-trained models do not have optimal architectures for the CIFAR data. For instance, the first layer of the pre-trained ResNet-50 model contains $7\times7$ filters, whereas a $3\times3$ filter is more suitable for the CIFAR data. On the other hand, state-of-the-art baseline models are usually initialized by the \emph{MSRA} method \cite{he2015delving}, while we use the \emph{Xavier} method \cite{glorot2010understanding} to remain consistent with the pre-trained models.

\begin{figure*}[!htbp]
\begin{center}
{\includegraphics[width=0.325\textwidth]{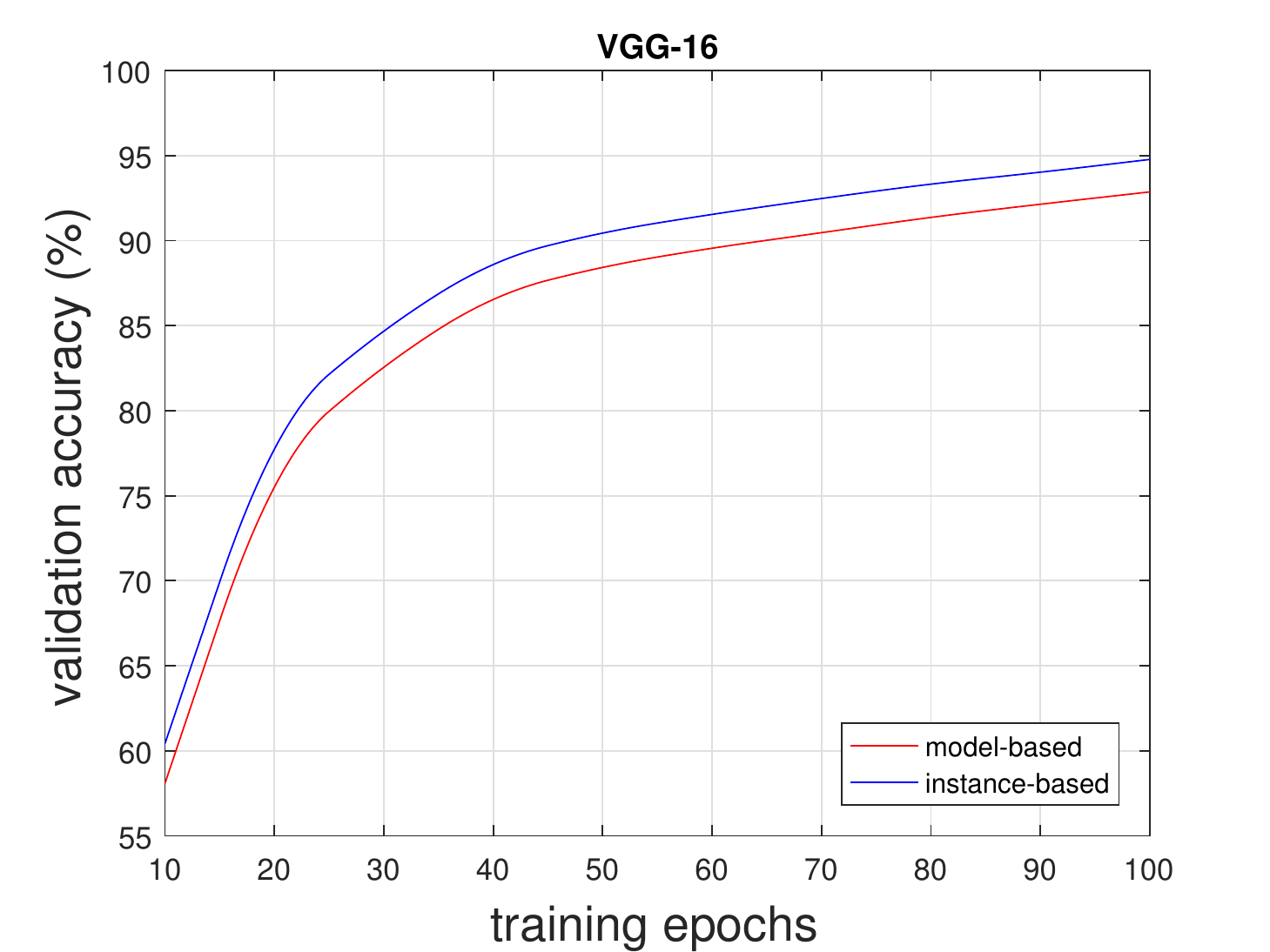}}
{\includegraphics[width=0.325\textwidth]{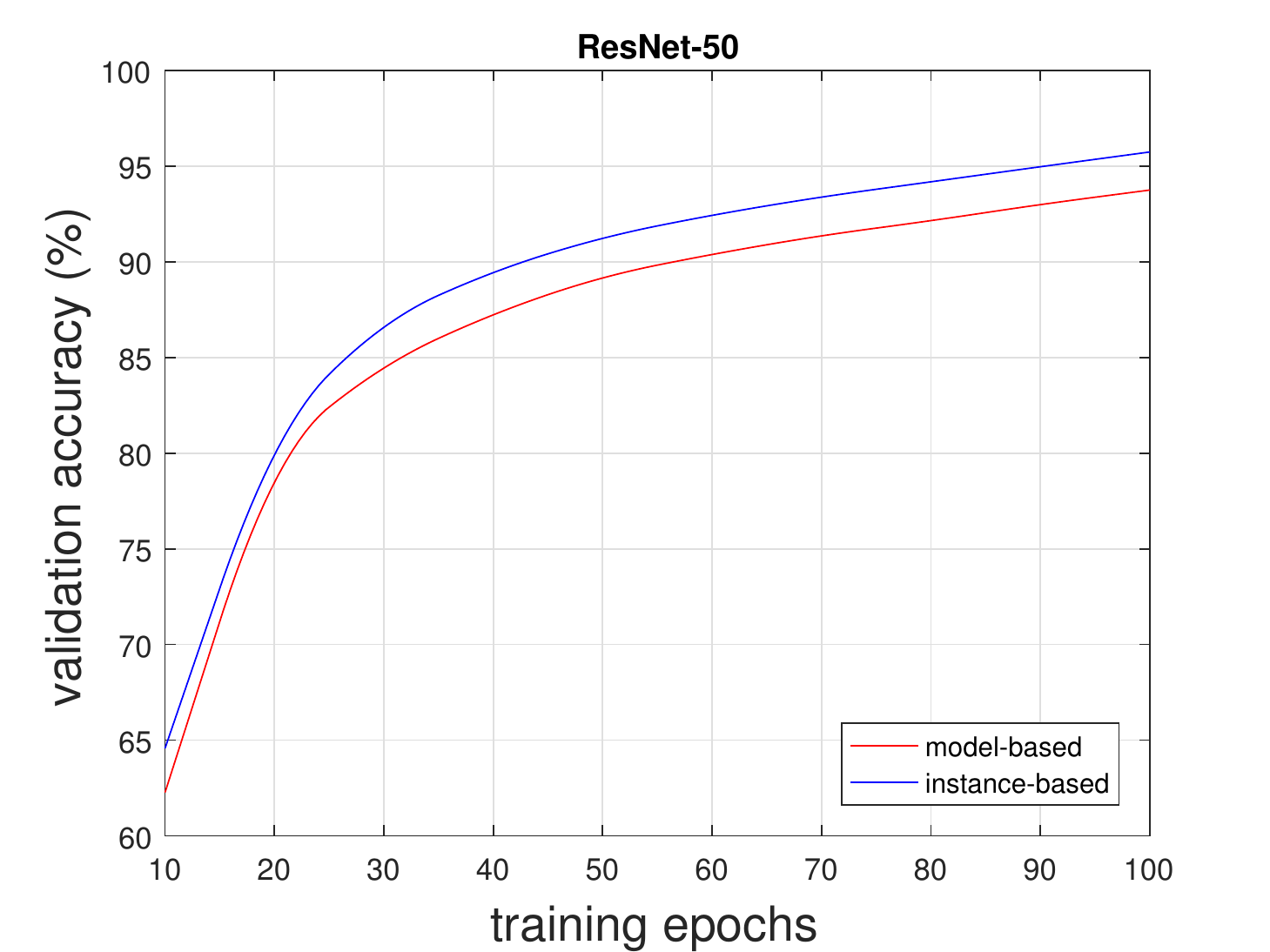}}
{\includegraphics[width=0.325\textwidth]{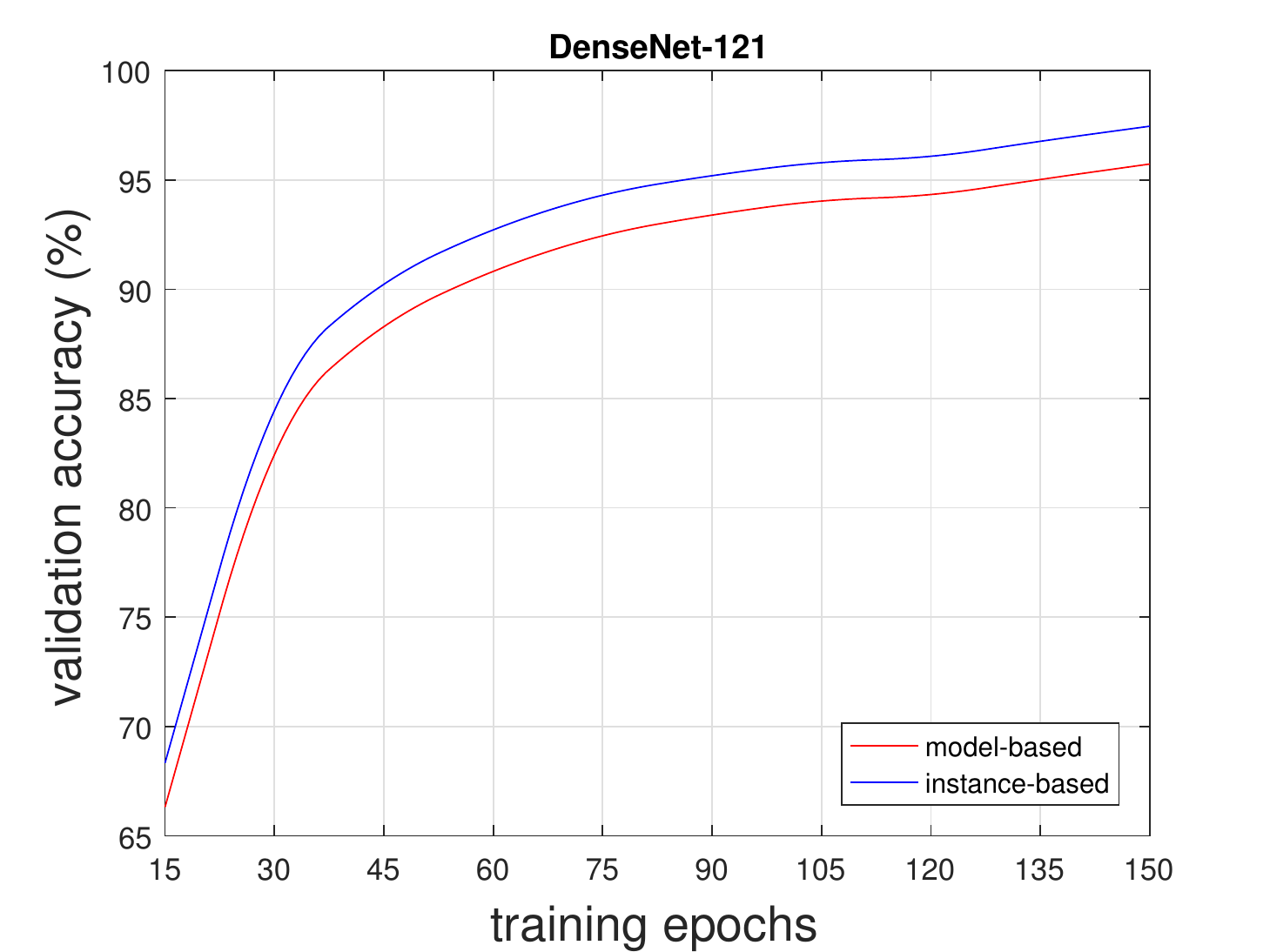}}

\caption{Curves of validation accuracy of each selected model on the CIFAR-10 classification. The values are recorded every 10 epochs for the VGG-16, the ResNet-50, and every 15 epochs for the DenseNet-121. \emph{Left:} curves of the VGG-16; \emph{Middle:} curves of the ResNet-50; \emph{Right:} curves of the DenseNet-121.}
\label{trainacc}
\end{center}
\vspace{-3mm}
\end{figure*}

In addition, we show the validation accuracy of the three models in Figure \ref{trainacc}. The value is recorded every 10 epochs for the VGG-16 and ResNet-50, whereas every 15 epochs for the DenseNet-121. As illustrated, fine-tuning with our approach consistently has higher validation accuracy during fine-tuning processes. This evidence indirectly proves that by optimizing the training data in a target domain, more significant features can be learned by deep learning models. Moreover, given a testing image which is misclassified by the model-based approach, the learned feature maps of the 23rd convolutional layer in the ResNet-50 are shown in Figure \ref{filters}. There are 1024 maps of this layer, we only show the first 64 for better page layout. It can be seen that, with the aid of our approach, the model can learn more important clues, such as the bright parts, that usually correspond to the edge information. The improvement can be attributed to the removal of disturbing features due to the training data optimization in the target domain. 

\renewcommand{\arraystretch}{1.06}
\setlength{\tabcolsep}{1.0em}
\begin{table}[]
\tiny
\resizebox{0.45\textwidth}{!}{%
\begin{tabular}{l|c|c}
\hline
\multicolumn{1}{c|}{Model} & Frozen Layers & CIFAR-10  \\ \hline

& 20 & 10.27 \\ 

ResNet-50 & 30 & 12.08 \\ 

& 40 & 13.98 \\ \hline
\end{tabular}%
}
\vspace{0.5ex}
\caption{Results of the ResNet-50 on the CIFAR-10 classification when fine-tuning with different frozen layers. The performance is measured in testing error rate (\%).}
\label{4-2-2table}
\vspace{-3mm}
\end{table}

\subsection{Loading Parameters for Shallow Layers}
\label{local}
The SVHN data is quite different from the ImageNet data, hence these two datasets do not share similar feature space. Hence, loading all pre-trained parameters is not appropriate for the SVHN classification. On the other hand, it is challenging to find a pre-trained model that was learned from a similar source domain. Therefore, we still exploit the ImageNet pre-trained models, but only load the parameters of their shallow layers.  

It is widely acknowledged that different images may still share similar global features. Therefore, for the two pre-trained models, we only preserve the shallow layers that learned global image features. Specifically, we create a new model which has the same architecture as the selected pre-trained model. We initialize the shallow layers by loading the pre-trained parameters, and randomly initialize the other layers by the \emph{Xavier} method \cite{glorot2010understanding}. Here, shallow layers refer to the first 10, and 31 layers of the ResNet-50, and DenseNet-121, respectively. In addition, the number of classes in fully connected layers is set to 10. Similar to the MNIST and CIFAR data, image dimension of the SVHN data is much smaller than that of the ImageNet data. Therefore, we need to load pre-trained parameters manually as discussed in section \ref{cifarclassify} to address the issue of incompatible dimension. A mini-batch size of 128 is used for the ResNet-50, and 64 for the DenseNet-121. We fine-tune the ResNet for 20 epochs and the DenseNet for 26 epochs. Learning rate is scheduled to reduce to 0.0001 at the 50\% of the total epochs. For training from scratch, the models are randomly initialized and both trained for 40 epochs.

\renewcommand{\arraystretch}{1.06}
\setlength{\tabcolsep}{1.0em}
\begin{table}[]
\tiny
\resizebox{0.45\textwidth}{!}{%
\begin{tabular}{c|l|c}
\hline
\multicolumn{1}{c|}{Strategy} & Model & SVHN \\ \hline
    
from Scratch & ResNet-50  & 2.12  \\ 
  & DenseNet-121 & 1.83  \\\hline

Model-based TL & ResNet-50  & 2.03  \\ 
  & DenseNet-121 & 1.72  \\\hline 

Instance-based TL & ResNet-50  & 1.35  \\ 
  & DenseNet-121 & 1.12   \\\hline
\end{tabular}%
}
\vspace{0.5ex}
\caption{Testing error rates (\%) of the selected models, which are trained by different approaches, on the SVHN classification.}
\label{4-3table}
\vspace{-3mm}
\end{table}

Similarly, we show three groups of results in Table \ref{4-3table}, corresponding to training from scratch, the model-based deep transfer learning, and our instance-based approach. As can be seen, our approach can still decrease the testing error rates, however, with a limited margin. This is partly because using the ImageNet pre-trained models to estimate the influence of each training sample in the SVHN dataset has bias, even though only shallow layers of the pre-trained models are transferred for fine-tuning.

\subsection{Non-Uniform Distributed Data}
In general, testing data usually follows uniform distribution or near uniform distribution. For instance, in the ILSVRC dataset, testing samples are near uniformly distributed in 1000 different classes. In the two CIFAR datasets, testing images are evenly distributed in 10 and 100 classes, respectively. Nevertheless, in real applications, such as in transductive problem, we know in advance that testing data may follow a non-uniform distribution. For example, in a testing set, 90\% of the samples are from only one class \emph{\textbf{A}}, and the rest of the samples are from other classes. In such a case, conventionally trained models cannot yield the best results. To improve performance, one intuitive option could be learning a new classifier by re-organizing the training data and adjusting the training strategy. For instance, changing the training data labels to have two classes only: \emph{\textbf{A}}, and \emph{\textbf{not A}}, and then training a binary classifier to solve the problem. However, this approach has two drawbacks: firstly, re-organizing training data is extremely time-consuming and even impractical; secondly, the testing data of the other classes cannot be accurately classified. Another option could be penalizing a loss function by using different weight values. During training, if a sample of the class \emph{\textbf{A}} is misclassified, a larger weight value can be used to penalize the loss. However, such a strategy lacks flexibility and highly relies on parameter-tuning skills.

In fact, our approach can effectively address this problem. 
When estimating the influence of the training data in a target domain, we can only remove the training samples which will lower the classification accuracy of the validation samples that belong to the class \emph{\textbf{A}}. To validate the effect, we construct two new testing sets based on the two CIFAR datasets. Specifically, for the CIFAR-10 dataset, we randomly choose a class, and randomly select 100 testing samples from it each time. We horizontally flip or rotate these images and save the resulting images as the new testing samples. We repeat this step for 91 times, hence it will generate 9100 new testing images belonging to the chosen class. Afterwards, for the remaining 9 classes, we randomly select 100 different testing images from each class, and save them in the new testing set, which will have 10000 images in total. Similarly, for the CIFAR-100 dataset, one class is randomly selected, and we randomly choose 10 testing samples from this class. Then we follow the same step and repeat it for 901 times to obtain 9010 new testing samples, and randomly select 10 different testing samples from each of the remaining classes. Hence the new testing set still contains 10000 images.

\renewcommand{\arraystretch}{1.06}
\setlength{\tabcolsep}{1.0em}
\begin{table}[]
\resizebox{0.45\textwidth}{!}{%
\begin{tabular}{c|l|c|c}
\hline
\multicolumn{1}{c|}{Strategy} & Model & CIFAR-10+ & CIFAR-100+ \\ \hline
from Scratch & ResNet-50  & 7.05 & 28.75 \\ 
  & DenseNet-121 & 5.02 & 21.97  \\\hline

Model-based TL & ResNet-50  & 6.35 & 28.12 \\ 
  & DenseNet-121 & 4.41 & 21.03  \\\hline 

Instance-based TL & ResNet-50  & 4.18 & 25.87 \\ 
  & DenseNet-121 & 2.80 & 19.26  \\\hline
\end{tabular}%
}
\vspace{0.5ex}
\caption{Testing error rates (\%) of the selected models, which are trained by different approaches, on the re-defined CIFAR classification. $+$ means the testing set is reconstructed.}
\label{4-4table}
\vspace{-3mm}
\end{table}

We follow the same training specifications as in section \ref{cifarclassify}, and compare the performance of the different training strategies in Table \ref{4-4table}. As illustrated, our instance-based approach can effectively handle such a non-uniform data distribution in the testing set, and give the best results. Moreover, to achieve this effect, we only need to make a minor change when applying the algorithm $\pmb{\mathcal{A}}$ to optimize the training set in the target domain. Specifically, when computing the influence value of each training sample, we only use the validation samples of class \emph{\textbf{A}} as a reference instead of considering all the validation samples. Therefore, our approach is flexible and simple to be deployed.

\subsection{Local Fine-Tuning for Little Data}
For all the experiments conducted above, there are sufficient training data in the target domain. In fact, a typical application of deep transfer learning is fine-tuning a pre-trained model with insufficient training data in a target domain. To validate our approach from this perspective, we conduct an experiment on a classical classification task: Cats vs Dogs.
There are 25000 training and 12500 validation images in the original dataset \cite{asirra-a-captcha}. However, to obtain a dataset which actually contains very few samples, we follow the steps in \cite{Cholletlittle} to select only 1000 images from the \textquoteleft Dog\textquoteright ~and \textquoteleft Cat\textquoteright ~class, respectively. Hence there are 2000 images for training. Similarly, 400 images are selected from each class for validation purpose. In addition, we choose 200 images from each class to build a testing set. All the images are chosen from the original training sets, and there is no overlapping. The images in this task are similar to the samples in the ImageNet dataset, we thus choose the ImageNet pre-trained models and fine-tune them on the new dataset.

\renewcommand{\arraystretch}{1.06}
\setlength{\tabcolsep}{1.0em}
\begin{table}[]
\tiny
\resizebox{0.45\textwidth}{!}{%
\begin{tabular}{c|l|c}
\hline
\multicolumn{1}{c|}{Strategy} & Model & Cats vs Dogs  \\ \hline

Model-based TL & ResNet-50  & 5.01  \\
  & DenseNet-121 & 4.64  \\\hline 

Instance-based TL & ResNet-50  & 3.38  \\ 
  & DenseNet-121 & 3.17  \\\hline
\end{tabular}%
}
\vspace{0.5ex}
\caption{Testing error rates (\%) of the selected models, which are trained by two approaches, on the problem of Cats vs Dogs.}
\label{4-5table}
\vspace{-3mm}
\end{table}

Since the training data is very limited, it is not appropriate to train a model from scratch due to the over-fitting risk. Therefore, we freeze all convolutional blocks of the selected pre-trained models and only fine-tune the fully connected layers. Prior to fine-tuning, we optimize the composed training set by following the algorithm $\pmb{\mathcal{A}}$. We replace the fully connected layers of the pre-trained models with randomly initialized fully connected layers, and set the number of classes to 2. The images are only augmented by horizontal flipping, zooming and shearing with a range of 0.2. We use a mini-batch size of 16 and fine-tune the ResNet-50 for 50 epochs and the DenseNet-121 for 60 epochs. The learning rate is reduced to 0.0001 at 50\% of the total epochs.  

Two groups of results from the model-based approach and our instance-based approach are compared in Table \ref{4-5table}.  Unlike the results discussed in section \ref{local}, our approach in this task improves the testing performance by a larger margin. This is partly because the parameters of the pre-trained models remain unchanged during the fine-tuning, hence the bias of the influence estimation is more or less mitigated.

\section{Conclusion}
\label{conclusion}

In this paper, we explore the feasibility and effectiveness of integrating instance-based transfer learning with deep learning. We propose a training scheme which exploits pre-trained models from a source domain to investigate the training data influence in a target domain. By removing the training samples that have negative impact, the training set in a target domain can be optimized. Unlike the model-based deep transfer learning, pre-trained models in our approach will be fine-tuned with an optimized training set, and the risk of learning some disturbing features in the models can be suppressed. Extensive experiments have demonstrated that our approach is compatible with the well-known pre-trained models on image classification problem.

\section*{Acknowledgement}

We gratefully acknowledge the support of NVIDIA Corporation with the donation of the Titan Xp and Titan X GPU used for this research to both Austin Peay State University and Montclair State University.

{\small
\bibliographystyle{ieee}
\bibliography{Reference}
}

\end{document}